\documentclass[10pt,twocolumn,letterpaper]{article}

\usepackage{ijcb}
\usepackage{times}
\usepackage{epsfig}
\usepackage{graphicx}
\usepackage{amsmath}
\usepackage{amssymb}
\usepackage{multirow}
\usepackage[table, svgnames]{xcolor}

\ijcbfinalcopy 

\ifijcbfinal\fi

\makeatletter
\def\ps@IEEEtitlepagestyle{
\def\@oddfoot{\mycopyrightnotice}
\def\@evenfoot{}
}
\def\mycopyrightnotice{
{\hfill \footnotesize 978-1-7281-9186-7/20/\$31.00 \copyright 2020 IEEE\hfill}
}
\makeatother

\begin{document}

\title{An Assessment of GANs for Identity-related Applications}

\author{\parbox{16cm}{\centering
    {\large Richard T. Marriott$^{1,2}$, Safa Madiouni$^2$, Sami Romdhani$^2$\\St\'ephane Gentric$^2$ and Liming Chen$^1$}\\
    {\normalsize
    $^1$ Ecole Centrale de Lyon, Ecully, France\\
    $^2$ IDEMIA, Courbevoie, France}}
}

\maketitle
\thispagestyle{empty}

\begin{abstract}
Generative Adversarial Networks (GANs) are now capable of producing synthetic face images of exceptionally high visual quality. In parallel to the development of GANs themselves, efforts have been made to develop metrics to objectively assess the characteristics of the synthetic images, mainly focusing on visual quality and the variety of images. Little work has been done, however, to assess overfitting of GANs and their ability to generate new identities. In this paper we apply a state of the art biometric network to various datasets of synthetic images and perform a thorough assessment of their identity-related characteristics. We conclude that GANs can indeed be used to generate new, imagined identities meaning that applications such as anonymisation of image sets and augmentation of training datasets with distractor images are viable applications. We also assess the ability of GANs to disentangle identity from other image characteristics and propose a novel GAN triplet loss that we show to improve this disentanglement.
\end{abstract}

\let\thefootnote\relax\footnotetext{\mycopyrightnotice}

\section{Introduction}

Generative Adversarial Networks (GANs) \cite{goodfellow2014generative} are now well known for their ability to generate highly realistic data-samples. Perhaps most famously, this includes high resolution images of faces such as those of the Progressive GAN \cite{karras2018progressive}. Whilst these synthetic images are of impressive quality, there is still debate as to their utility beyond purely aesthetic applications such as image in-painting \cite{yu2019free} and image editing \cite{shen2020interpreting}. For example, with the recent implementation of the General Data Protection Regulation (GDPR) in Europe, a natural question to ask is whether GANs can be used to generate sets of synthetic identities as a method of dataset anonymisation. Many researchers have also attempted to use GAN-generated data to augment training datasets with varying degrees of success \cite{liu2019generative, trigueros2018generating, zheng2017unlabeled}.

\begin{figure}[t]
  \centering
  \includegraphics[width=0.2\linewidth]{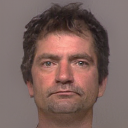}\includegraphics[width=0.2\linewidth]{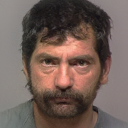}\includegraphics[width=0.2\linewidth]{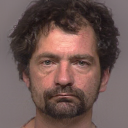}\includegraphics[width=0.2\linewidth]{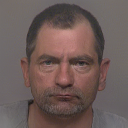}\includegraphics[width=0.2\linewidth]{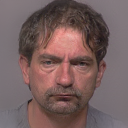}
  \includegraphics[width=0.2\linewidth]{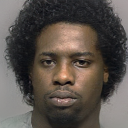}\includegraphics[width=0.2\linewidth]{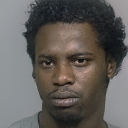}\includegraphics[width=0.2\linewidth]{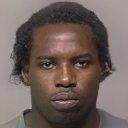}\includegraphics[width=0.2\linewidth]{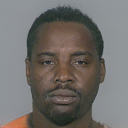}\includegraphics[width=0.2\linewidth]{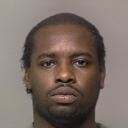}
  \includegraphics[width=0.2\linewidth]{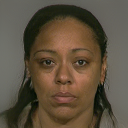}\includegraphics[width=0.2\linewidth]{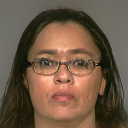}\includegraphics[width=0.2\linewidth]{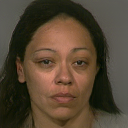}\includegraphics[width=0.2\linewidth]{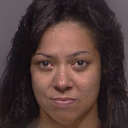}\includegraphics[width=0.2\linewidth]{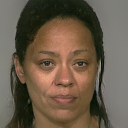}
  \includegraphics[width=0.2\linewidth]{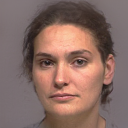}\includegraphics[width=0.2\linewidth]{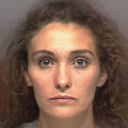}\includegraphics[width=0.2\linewidth]{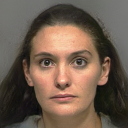}\includegraphics[width=0.2\linewidth]{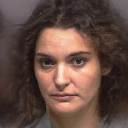}\includegraphics[width=0.2\linewidth]{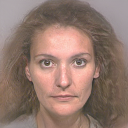}
  \caption{Synthetic samples generated by an SD-GAN \cite{donahue2018semantically} trained on a proprietary dataset of mugshots.}
  \label{fig:B61_samples}
\end{figure}

In this paper we study the ability of GANs to generate new synthetic identities, and their ability to disentangle identity from the description of other image characteristics in the latent space. Our investigations show that the former is indeed possible whereas the latter is problematic. Our contributions are:


\begin{enumerate}
   \item An explicit demonstration that GANs do indeed generate new identities;
   \item Introduction of biometric analysis as a measure the degree of overfitting and mode-collapse in GANs;
   \item A novel GAN triplet loss that improves disentanglement of identity in SD-GANs \cite{donahue2018semantically}.
\end{enumerate}

The paper is organised as follows: in Section \ref{Sec. Related work} we present other works that attempt to quantify the characteristics of GAN-generated images as well as works attempting to disentangle identity; in Section \ref{Sec. GAN_Applications} we discuss potential applications of GAN-generated images; in Section \ref{Sec. Dynamics} we assess overfitting in GANs with respect to identity; in Section \ref{Sec. Disentanglement} we introduce our novel GAN triplet loss and compare its ability to disentangle identity with various other methods; and in Section \ref{Sec. Conclusions}, we draw conclusions as to legitimate uses of GAN-generated data.

\section{Related work}\label{Sec. Related work}

\subsection{GAN metrics}\label{Sec. RW_GAN_metrics}



The \textit{de facto} standards for assessing GANs are currently the Inception Score (IS) \cite{salimans2016improved}, the Fr\'echet Inception Distance (FID) \cite{heusel2017gans}, and Precision and Recall (P\&R) \cite{sajjadi2018assessing, kynkaanniemi2019improved}. IS and FID each combine measures of the quality and variety of generated images whereas P\&R avoids ambiguity by providing Precision as a measure of image quality and Recall as a measure of variety. None of these metrics, however, is sensitive to the problem of overfitting. A ``Memory GAN'' that memorises and reproduces exact copies of training data would still give optimal IS, FID and P\&R scores.

The lack of a suitable metric for assessing overfitting was identified in \cite{lucic2018gans} although, despite being stated as one of the motivations for proposing their measure of P\&R calculated for toy datasets, the paper does not elaborate on how overfitting might be measured. In \cite{webster2019detecting}, overfitting is measured by analysing discrepancies in the distributions of reconstruction error for training and validation images upon inversion of the generator. The method appears to be informative. However, in practice it is expensive to invert the GAN's generator for all training images. In this work, we efficiently project datasets of facial images into a biometric feature-space where comparisons can be made based on subtle yet robust features. Doing so is primarily motivated by our interest in identity. However, biometric datasets may serve as a useful standard for evaluating GANs in general.

\subsection{Disentanglement}\label{Sec. RW_Disentanglement}

The second analysis presented in this paper is of the ability of GANs to disentangle identity from other image characteristics in their latent spaces. Such an ability would enable, for example, augmentation of datasets with mated sets of images, as opposed to augmentation with distractor images of arbitrary identity. In the literature we find three general methods of controlling images generated by GANs:

\begin{enumerate}
	\item Training a conditional GAN (cGAN) and adding a biometric constraint on the identity. In this work we evaluate IVI-GAN \cite{marriott2020taking}. However, there are many other works that fall into this category \cite{tran2018representation, trigueros2018generating}.
	\item Training a standard GAN and retrospectively discovering semantically meaningful axes of variation in the latent space \cite{shen2020interpreting, harkonen2020ganspace}. We assess the InterFaceGAN method of \cite{shen2020interpreting}. The method does not explicitly avoid changes to identity upon traversing the GAN's latent space and assumes that a robustly trained GAN will naturally learn to disentangle semantic factors.
	\item Training the discriminator of the GAN to act upon pairs of images and to simultaneously penalise poor realism and poor identity consistency. The only work of which we are aware that has previously attempted this is that of \cite{donahue2018semantically}.
\end{enumerate}

Examples of each of these types of method are described in more detail and evaluated in Section \ref{Sec. Disentanglement}.

\section{GAN Applications}\label{Sec. GAN_Applications}

\subsection{Data-anonymisation}\label{Sec. Applications_Data_anonymisation}

The anonymisation of data is increasingly becoming a research topic of interest. Unrestricted use of datasets of face-images depends on the ability to remove information that allows identification of the original subjects. A crude method of doing this might be simply to detect faces  and set those pixels to zero. Ideally, however, we would like to remove identity information without affecting other semantic properties of the image or of the dataset as a whole.

Most methods in the literature tackle the problem of data-anonymisation by modifying individual images, leaving all properties untouched except for the identity. Some of the more successful early attempts at doing so involved reconstructing the subject of the original image using a 3D morphable model (3DMM) \cite{blanz1999morphable} before modifying the model's shape parameters to change the identity \cite{gross2008semi, samarzija2014approach}. In \cite{meden2017face}, a CNN conditioned on biometric vectors is used to generate a face-image that is a mixture of nearby identities before blending it back into the original image. More recent methods treat the problem as image-to-image translation or as image in-painting. In \cite{wu2018privacy} an auto-encoder is trained to optimise a reconstruction loss, but also a biometric loss to enforce distance between the original and translated images in an identity feature-space. In \cite{sun2018natural} and \cite{hukkelaas2019deepprivacy} faces are first obscured and then auto-encoders trained to in-paint the obscured regions, harmonising them as best they can with the rest of the image in order to fool an adversarial loss. In \cite{sun2018hybrid} a hybrid 3DMM plus in-painting method is used.

Due to the constraint of having to maintain the specific contexts of individual images, the aforementioned problem is difficult and generated images tend to be of low quality, either lacking in high-frequency detail or containing obvious artefacts. It is possible, however, to generate images of the required quality by training a GAN whose soul objective is to accurately approximate the distribution of real images as a whole, i.e. not endeavouring to preserve the specific contexts of individual images. As mentioned in Section \ref{Sec. RW_GAN_metrics}, although much work has been done to assess the quality and variety of GAN-generated images, there is no work explicitly demonstrating GANs to consistently generate new identities.


\subsection{Data-augmentation}\label{Sec. Applications_Data_augmentation}

Data-augmentation using synthetic identities might take one of two forms:

\begin{enumerate}
	\item Supervised: generation of sets of ID-labelled images where each set contains a ``unique'' identity;
	\item Semi-supervised: generation of examples of arbitrary identity (the goal being to avoid classifying them as the labelled identities of the training dataset).
\end{enumerate}

We are aware of only a handful of examples of ``supervised'' data-augmentation using synthetic identities. The works of \cite{trigueros2018generating}, \cite{gecer2018semi} and \cite{gecer2019synthesizing} each train facial recognition (FR) networks using both real and synthetic identities simultaneously. This makes it important to be sure that new, synthetic identities are being generated to ensure that collisions with existing identities are rare. If this were not the case, there would be undesirable consequences for the compactness of learned class-representations in the feature-space.

To our knowledge, there are no evaluations of semi-supervised learning for FR in the literature. However, the technique has been evaluated for the similar problem of person re-identification. Semi-supervised learning using GANs was originally proposed in \cite{salimans2016improved}, and concurrently in \cite{odena2016semi}. These works proposed what we will term a ``$k+1$'' method in which the classifier that they are trying to improve (which has $k$ classes) is also trained to be the discriminator of a GAN by assigning synthetic images to an additional, $k+1^{th}$ class. Why this was found to improve classification of real images is not entirely clear. One might imagine that it has something to do with multi-task learning, or alternatively that the larger generator+classifier network acts as teacher of the smaller, student network that is the classifier alone.


The findings of \cite{dai2017good} suggest that it might not be straightforward to train a classifier with a $k+1^{th}$ class using realistic face images generated by a standard GAN. The assumption that the GAN generates \textit{only} new identities might be too strong. An alternative semi-supervised training method that makes a weaker assumption is Label Smoothing Regularisation for Outliers (LSRO) \cite{zheng2017unlabeled}. LSRO takes a randomly generated set of synthetic images and assigns each a label vector with uniformly distributed values, as opposed to a one-hot vector indicating a specific class. This label-smoothing for the synthetic images is thought to have a regularising effect on the classifier, reducing the classifier's confidence when seeing identities that are similar to those in the training set.

For both $k+1$ methods and LSRO, it is important that the synthetic training set contains identities different from those in the training dataset. In the following section we show results that explicitly confirm this to be the case.

\section{Overfitting of identity}\label{Sec. Dynamics}

As previously discussed, existing GAN metrics (e.g. IS, FID and P\&R) measure the quality and variety of generated images but do not indicate whether overfitting to the training dataset has occurred. This means that these metrics give no indication of whether generated images depict new subjects or whether identities are biased towards those of the training dataset. In this section, we explicitly assess the proximity of generated images to faces in the training dataset using a state-of-the-art biometric network based on \cite{deng2019arcface}. Inevitably, some images match more closely than others. There is no threshold that we can define, however, that tells us when any individual image has ``overfit'' to the training dataset. We wish for the GAN to approximate the real-world distribution of images of faces and, as in the real world, sometimes lookalikes do occur. Instead, we must ensure that the frequency at which lookalikes occur is not significantly greater than in the training dataset. We therefore make a comparison between two distributions of matching scores: the distribution of matching scores between synthetic images and training images, and the distribution of matching scores between the training images themselves.

The matching scores used in our analysis are analogous to cosine similarities in the feature space of the biometric network. Feature vectors were generated and then converted to scores such that higher scores represent stronger similarity between identities. In the following subsection we present details and results of this study, showing that new identities are indeed generated by GANs; then in Section \ref{Sec. Measuring mode collapse} we describe how a similar assessment of matching scores can be used as a measure of mode-collapse.

\subsection{Overfitting results}\label{Sec. Generation of new identities}

\begin{figure*}[t]
  \centering
  \includegraphics[width=0.5\linewidth]{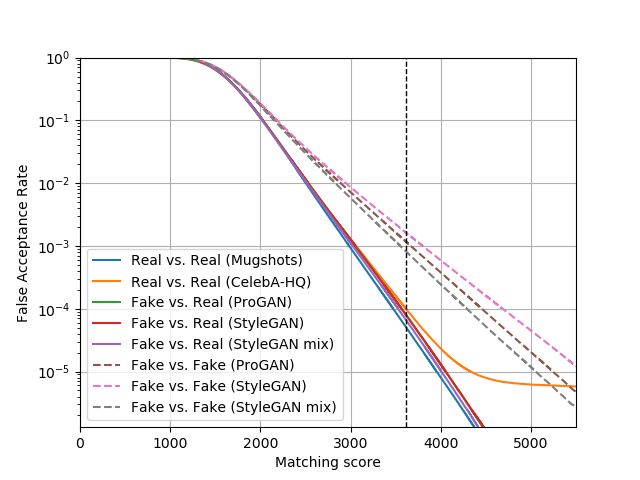}\includegraphics[width=0.5\linewidth]{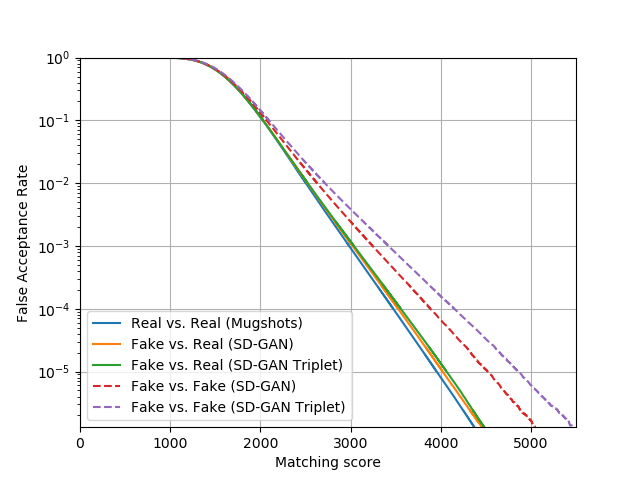}
  \caption{False acceptance rates across biometric matching score thresholds for \textit{all} pairs of non-mated images either within or between real and synthetic datasets as indicated in the legend.}
  \label{fig:CelebA_HQ_FARs_comp}
\end{figure*}

\begin{figure*}[t]
  \centering
  \includegraphics[width=0.5\linewidth]{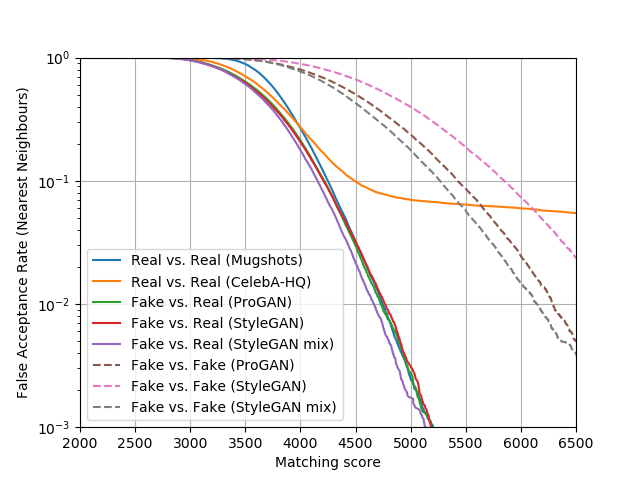}\includegraphics[width=0.5\linewidth]{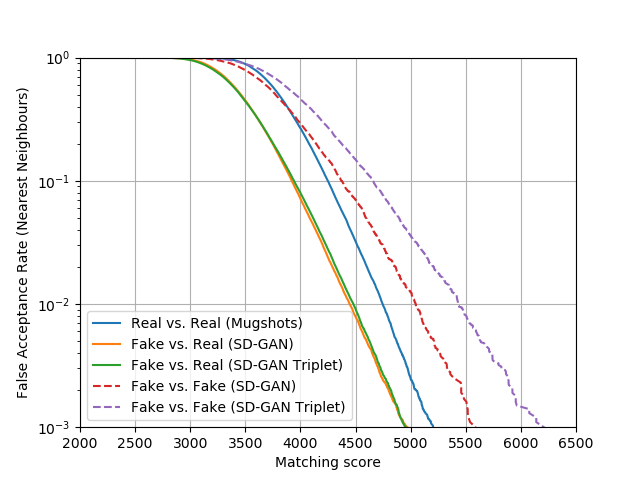}
  \caption{False acceptance rates of \textit{nearest neighbour} images across biometric matching score thresholds for non-mated image-pairs either within or between real and synthetic datasets as indicated in the legend.}
  \label{fig:CelebA_HQ_FARs_NNs_comp}
\end{figure*}

In Figures \ref{fig:CelebA_HQ_FARs_comp} (left) and \ref{fig:CelebA_HQ_FARs_NNs_comp} (left) we assess overfitting of the official versions of the Progressive GAN \cite{karras2018progressive}, and of the more recent StyleGAN \cite{karras2019style}. We show results for two versions of StyleGAN: the default, validation version, and a version with StyleGAN's style-mixing enabled, labelled as ``mix'' in the figures and in Table \ref{tab:FARs}. Style-mixing is typically only enabled during training of StyleGAN to help disentangle different scales. However, we note that reducing the variability of input to StyleGAN by disabling style-mixing causes increased mode-collapse. Each GAN was trained on the CelebA-HQ dataset containing $30,000$ images at $1024\times 1024$ resolution. Figures \ref{fig:CelebA_HQ_FARs_comp} (right) and \ref{fig:CelebA_HQ_FARs_NNs_comp} (right) show the same analysis for two brands of SD-GAN \cite{donahue2018semantically} trained on a proprietary dataset of mugshot images. SD-GAN are capable of disentangling identity from other forms of variation and will be described in Section \ref{Sec. Disentanglement} where we also introduce our novel GAN triplet loss. For the Progressive GAN and StyleGAN, datasets of $30,000$ images of random, synthetic identities were generated, and for the SD-GAN, $10,000$ images for $1000$ identities ($10$ images each).

Figure \ref{fig:CelebA_HQ_FARs_comp} (left) shows FAR as a function of matching score threshold. The curves labelled ``Real vs. Real'' show the distributions within the proprietary ``Mugshots'' dataset, and within CelebA-HQ. It can be seen that, within CelebA-HQ, the proportion of non-mated identities found to match is relatively high, even for large matching score thresholds of $5,000$ and above. Labelling in CelebA (and therefore in CelebA-HQ) is noisy, with many subjects belonging to more than one ID category. This causes the level of similarity between identities in the real data to be overestimated. Despite this problem, we have left the curve as an example of the dynamics that can be expected if duplicate identities are present. The ``Mugshots'' curve has been included as an example of the dynamics one should expect of a cleanly labelled FR dataset.

The three curves labelled as ``Fake vs. Real'' show the distributions of matching scores between the dataset generated by the indicated GAN and CelebA-HQ. (Note that the green curve of the Progressive GAN is hidden beneath the red curve of the standard StyleGAN.) Each of these curves demonstrates consistently lower matching frequencies than for the images of CelebA-HQ. What is more, the matching frequencies are in close agreement with those of the clean, Mugshots dataset indicating that the synthetic datasets do not resemble CelebA-HQ significantly more than you would expect non-mated identities to resemble one-another in a dataset of real images. Figure \ref{fig:CelebA_HQ_FARs_comp} (right) shows that the same is true of images generated by both flavours of SD-GAN. In this case, the dataset of mugshots was used for training of the GANs. The fact that Fake-Real matching scores are not significantly stronger than Real-Real scores means that GANs can be used to effectively anonymise datasets. We are therefore able to publish the synthetic images in Figures \ref{fig:B61_samples} and \ref{fig:Triplet_samples} without betraying the identities of the training dataset.

\begin{table}
\begin{center}
\resizebox{\columnwidth}{!}{%
\begin{tabular}{|l|cc|}
\hline
\multirow{2}{*}{Comparison} & \multicolumn{2}{c|}{False Acceptance Rate} \\
 & @3000 & @3614.5 \\
\hline\hline
Real vs. Real (CelebA-HQ)    & $1.29\times10^{-3}$ & $1.00\times10^{-4}$ \\
Fake vs. Real (ProGAN)       & $1.26\times10^{-3}$ & $0.80\times10^{-4}$ \\
Fake vs. Real (StyleGAN)     & $1.25\times10^{-3}$ & $0.80\times10^{-4}$ \\
Fake vs. Real (StyleGAN mix) & $1.11\times10^{-3}$ & $0.66\times10^{-4}$ \\
Fake vs. Fake (ProGAN)       & $7.19\times10^{-3}$ & $1.16\times10^{-3}$ \\
Fake vs. Fake (StyleGAN)     & $8.64\times10^{-3}$ & $1.62\times10^{-3}$ \\
Fake vs. Fake (StyleGAN mix) & $5.80\times10^{-3}$ & $0.82\times10^{-3}$ \\
\hline
\end{tabular}
}
\end{center}
\caption{FARs read from Figure \ref{fig:CelebA_HQ_FARs_comp} (left) at two thresholds.}
\label{tab:FARs}
\end{table}

Figure \ref{fig:CelebA_HQ_FARs_comp} shows distributions of matching scores for all image combinations, even those that map to distant parts of the identity feature-space. An alternative way to assess overfitting is to look only at the largest score for each image, i.e. to observe the distributions of nearest neighbour scores. These are plotted in Figure \ref{fig:CelebA_HQ_FARs_NNs_comp}. From these results we notice that the story is essentially the same: each of the ``Fake vs. Real'' curves displays matching scores that are either lower than or close to those found within the training datasets. This supports the conclusion that the GANs' generators are not overfitting to the training datasets.

\subsection{Mode-collapse of identity}\label{Sec. Measuring mode collapse}

In this section we discuss interpretation of the ``Fake vs. Fake'' curves (dashed lines) of Figures \ref{fig:CelebA_HQ_FARs_comp} and \ref{fig:CelebA_HQ_FARs_NNs_comp}. These curves represent the distributions of matching scores \textit{within} the synthetic datasets and show much higher frequencies of strong matches than are found within or between the other datasets (with the exception of CelebA-HQ at certain thresholds). This indicates that, although synthetic subjects do not strongly resemble those of the training dataset, they do strongly resemble one-another. This is a symptom of the well-known problem of mode-collapse in GANs in which well-separated random input vectors are mapped to similar points in image-space.

Similar to the case for overfitting, since the Progressive GAN and StyleGAN have no reason to treat identity features differently from any other image-feature, we can use the increase in ``Fake vs. Fake'' matching frequency as a general measure of mode-collapse in GANs. One might try to quantify the degree of mode-collapse as a single value by measuring the number of distinct identities being generated and representing this as a fraction of that for the dataset of real images. For example, the threshold of $3614.5$ was chosen to give an FAR of $1 \times 10^{-4}$ for CelebA-HQ. This means that the algorithm, in conjunction with this threshold, is capable of distinguishing $1/1\times10^{-4} = 10,000$ real identities. When applied to a synthetic dataset (say ``StyleGAN mix''), the biometric network finds only $1/0.82\times10^{-3} = 1220$ distinguishable identities, implying that the synthetic identities have collapsed to span a region of only $1220/10,000 = 12.2\%$ of that of the real distribution. However, by performing the equivalent calculation using values from Table \ref{tab:FARs} for the cruder threshold of $3000$, we find that the estimate of the level of mode-collapse improves to $1.29/5.80 = 22.2\%$; i.e. the estimation of the degree of mode-collapse is dependent on the precision with which data-points are represented in the feature space. To avoid this complication, we therefore recommend analysing the full dynamics of the FAR as presented in Figure \ref{fig:CelebA_HQ_FARs_comp} (or alternatively Figure \ref{fig:CelebA_HQ_FARs_NNs_comp}).

\section{Disentanglement of identity}\label{Sec. Disentanglement}

In order for GANs to be useful for supervised data-augmentation for FR, they must be able to adequately disentangle identity from other image properties. To assess this ability, we analyse distributions of matching scores within mated image sets intended to depict the same subject. If disentanglement is successful, we should expect the distribution of matching scores to be similar to that for sets of real images. We perform this analysis for datasets generated by an example of each type of GAN identified in Section \ref{Sec. RW_Disentanglement}. These methods are IVI-GAN \cite{marriott2020taking}, InterFaceGAN \cite{shen2020interpreting}, SD-GAN \cite{donahue2018semantically}, and also an SD-GAN trained using our novel GAN triplet loss that will be introduced in the following subsection. IVI-GAN and the Progressive GAN used by the InterFaceGAN method were trained using CelebA and CelebA-HQ respectively. The SD-GAN were trained on our proprietary dataset of mugshots since the discriminators must also learn reliable biometric functions. (CelebA is not cleanly labelled and contains fewer identities than are typically used to train state-of-the-art biometric networks such as that used by IVI-GAN.) Since pose is disentangled by both IVI-GAN and InterFaceGAN, we ensure a fair comparison with the SD-GAN by selecting pose parameters such that the standard deviation of yaw angles detected in generated images matches that of images generated by the SD-GAN. Each GAN was used to generate sets of ten images for 1000 identities as described in the subsections below.

\subsection{A Triplet Loss for SD-GANs}\label{Sec. Triplet loss}

\begin{figure}[t]
  \centering
  \includegraphics[width=0.6\linewidth]{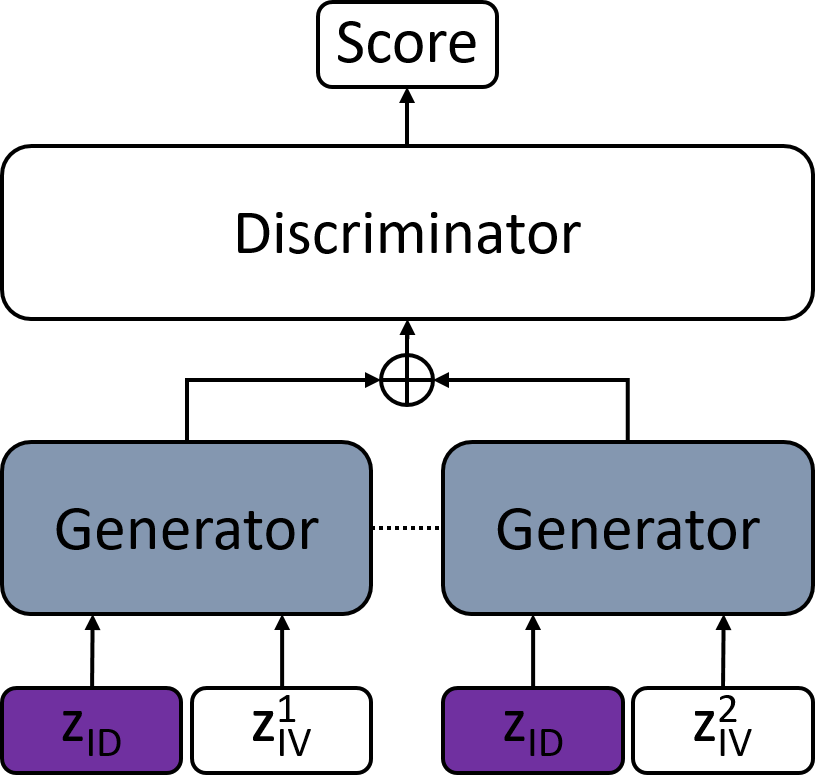}
  \caption{The SD-GAN architecture used in our evaluation. $\mathbf{z}_{ID}$, $\mathbf{z}_{IV}^1$ and $\mathbf{z}_{IV}^2$ are random vectors; the dotted line indicates shared network weights and the plus indicates channel-wise concatenation of generated images. (IV refers to ``Intra-class Variation''.)}
  \label{fig:SD-GAN_diagram}
\end{figure}

The triplet loss was first used for FR in \cite{schroff2015facenet}, derived from \cite{weinberger2009distance}, and is typically used to train classifier networks. It has two terms that each ensure desirable characteristics of embeddings in the discriminative feature space: a first term minimises the distance between pairs of images of the same class and ensures intra-class compactness, while a second term maximises the distance from one image of each pair to an image of a different class and ensures good inter-class separation. During training of an SD-GAN, in addition to learning to judge the realism of images, the discriminator performs a biometric function, learning to judge whether pairs of images share the same identity or not. Learning of this function is based on the consumption of only positive examples of matching identities. This could be thought of as being analogous to the first term of the original triplet loss that ensures good intra-class compactness. Our results show that the biometric function of the SD-GAN's discriminator can also benefit from the integration of a term that acts upon non-mated pairs of images.

We incorporate our triplet loss into an SD-GAN with Wasserstein loss \cite{pmlr-v70-arjovsky17a}. However, the same modification could be made to other GAN losses. The original Wasserstein loss functions for the discriminator and the generator can be written as follows:
\begin{align}
    \mathcal{L}_{\theta_D} &= \mathbb{E}_{\mathbf{x}\sim p_{d}}[D(\mathbf{x}; \theta_D)] - \mathbb{E}_{\mathbf{z}\sim\mathcal{N}}[D(G(\mathbf{z}; \theta_G); \theta_D)]\\
    \mathcal{L}_{\theta_G} &= \mathbb{E}_{\mathbf{z}\sim\mathcal{N}}[D(G(\mathbf{z}; \theta_G); \theta_D)]
\end{align}
where $\mathbf{x}$ are images selected at random from the real data distribution, $p_{d}$; $\mathbf{z}$ is a random vector whose values are selected randomly from a standard Gaussian distribution; and $\theta_D$ and $\theta_G$ parameterise the discriminator, $D$, and the generator, $G$, respectively. The equivalent loss functions for a standard SD-GAN are then written as
\begin{align}
	\mathcal{L}_{\theta_D} = &D(\mathbf{x}) - D(G(\mathbf{z}_{1}), G(\mathbf{z}_{2}))\\
	\mathcal{L}_{\theta_G} = &D(G(\mathbf{z}_{1}), G(\mathbf{z}_{2}))
\end{align}
where we have dropped the expectations and network parameters for simplicity of notation. Here, $\mathbf{x}$ is now a \textit{pair} of images of matching identity, and $\mathbf{z_1} = \left[ \mathbf{z}_{ID}, \mathbf{z}_{IV}^1 \right]$ and $\mathbf{z_2} = \left[ \mathbf{z}_{ID}, \mathbf{z}_{IV}^2 \right]$ are random vectors sharing the common subvector $\mathbf{z}_{ID}$. To integrate the ``imposter'' term of our GAN triplet loss, the losses are then modified to
\begin{align}
	\mathcal{L}_{\theta_D} = \label{Eq. Triplet D} &D(\mathbf{x}) - \frac{1}{2}[D(G(\mathbf{z}_{1}), G(\mathbf{z}_{2})) + D(\mathbf{\bar{x}})]\\
	&+ \lambda[D(G(\mathbf{z}_{1}), G(\mathbf{z}_{2})) - D(\mathbf{\bar{x}})]^2\nonumber\\
	\mathcal{L}_{\theta_G} = &D(G(\mathbf{z}_{1}), G(\mathbf{z}_{2}))
\end{align}
where $\mathbf{\bar{x}}$ is a pair of images containing non-mated identities. To strictly follow the analogy with the original triplet loss, one of these images would be shared with $\mathbf{x}$ and the second would be an imposter. However, in practice we sample $\mathbf{\bar{x}}$ randomly from a set of pre-defined image pairs that demonstrate high matching scores as judged by a biometric network.

The core idea of the GAN triplet loss is encapsulated in the second term of the discriminator loss: the discriminator is applied to the non-mated image pairs and the resulting scores averaged with those of the synthetic images to create a new ``fake data'' term. These two different ways of image pairs appearing to be ``fake'' (either being synthetic, or real but non-mated) are then contrasted with a real pair of matching images in the first term of the loss. Having added the term $D(\mathbf{\bar{x}})$, $\mathcal{L}_{\theta_D}$ could now be minimised by simply forcing apart the embeddings of $\mathbf{x}$ and $\mathbf{\bar{x}}$ in the feature space of the discriminator and ignoring the synthetic images. To avoid this, we add the third, quadratic term to ensure that the synthetic and non-matching terms retain roughly the same magnitude. This ensures that useful gradients are back-propagated to the generator. In our experiments we found that a weight of $\lambda = 0.001$ worked well and that a value of $\lambda = 0$ resulted in decreased image quality.

\begin{figure}[t]
  \centering
  \includegraphics[width=0.2\linewidth]{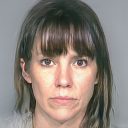}\includegraphics[width=0.2\linewidth]{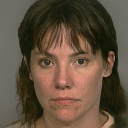}\includegraphics[width=0.2\linewidth]{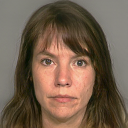}\includegraphics[width=0.2\linewidth]{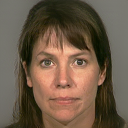}\includegraphics[width=0.2\linewidth]{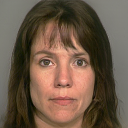}
  \includegraphics[width=0.2\linewidth]{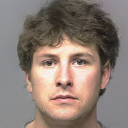}\includegraphics[width=0.2\linewidth]{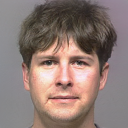}\includegraphics[width=0.2\linewidth]{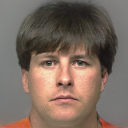}\includegraphics[width=0.2\linewidth]{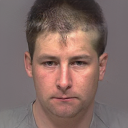}\includegraphics[width=0.2\linewidth]{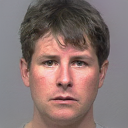}
  \caption{Synthetic samples generated by an SD-GAN trained using our GAN triplet loss.}
  \label{fig:Triplet_samples}
\end{figure}

As official code is not available for SD-GAN, we implemented our own version, building upon NVidia's Progressive GAN. A variety of formulations of SD-GAN is proposed in \cite{donahue2018semantically}. The one we opted for is depicted in Figure \ref{fig:SD-GAN_diagram}. Since the discriminator operates on pairs of images, we double the number of filters in each convolutional layer. We found that doing so improved the visual quality of generated images. Other than this modification, the architectures of the generator and discriminator are the same as those found in \cite{karras2018progressive}. Examples of synthetic image sets generated by SD-GAN and our SD-GAN with triplet loss can be found in Figures \ref{fig:B61_samples} and \ref{fig:Triplet_samples} respectively.

\subsection{IVI-GAN}\label{Sec. IVI-GAN}

\begin{figure}[t]
  \centering
  \includegraphics[width=0.2\linewidth]{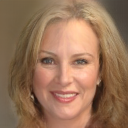}\includegraphics[width=0.2\linewidth]{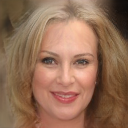}\includegraphics[width=0.2\linewidth]{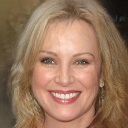}\includegraphics[width=0.2\linewidth]{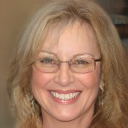}\includegraphics[width=0.2\linewidth]{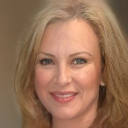}
  \includegraphics[width=0.2\linewidth]{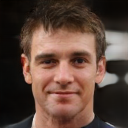}\includegraphics[width=0.2\linewidth]{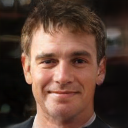}\includegraphics[width=0.2\linewidth]{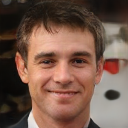}\includegraphics[width=0.2\linewidth]{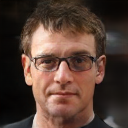}\includegraphics[width=0.2\linewidth]{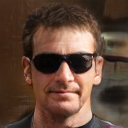}
  \caption{Two sets of synthetic samples generated by IVI-GAN \cite{marriott2020taking} with random pose, expression, lighting and eyewear.}
  \label{fig:IVI_samples}
\end{figure}

IVI-GAN \cite{marriott2020taking} is a conditional GAN trained to have disentangled parameters. Synthetic images, $\mathbf{x}$, are generated by a function $\mathbf{x} = G(\mathbf{z}, \mathbf{\rho}; \theta_G)$ where $\mathbf{z}$ is a vector of random values selected from a Gaussian distribution; $\mathbf{\rho}$ is a parameter vector consisting of vectors $\left[\mathbf{p}_1 ... \mathbf{p}_N\right]$, each with random values describing a different image attribute, for example, lighting or pose; and $\theta_G$ are the trained parameters of the Generator, $G$. A biometric constraint is applied to the generated images during training to ensure that $B(G(\mathbf{z}, \mathbf{\rho_1}; \theta_G); \theta_B) \approx B(G(\mathbf{z}, \mathbf{\rho_2}; \theta_G); \theta_B)$ where $\theta_B$ are the parameters of a pre-trained biometric network, $B$. This constraint ensures that the vector $\mathbf{z}$ contains information describing the generated identity and that the identity does not change upon varying $\mathbf{\rho}$. To generate our 1000 sets of images, 1000 random $\mathbf{z}$ were selected as well as ten random $\mathbf{\rho}$ for each, corresponding to varying pose, expression, lighting and eyewear. All parameters were selected as during training with the exception of the pose parameters which were selected from a standard Gaussian distribution but then scaled by 0.21 in order for the standard deviation of detected yaw angles to match those generated by SD-GAN. Two sets of samples generated by IVI-GAN can be seen in Figure \ref{fig:IVI_samples}.

\begin{figure}[t]
  \centering
  \includegraphics[width=0.2\linewidth]{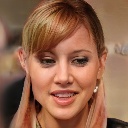}\includegraphics[width=0.2\linewidth]{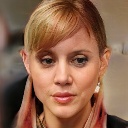}\includegraphics[width=0.2\linewidth]{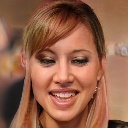}\includegraphics[width=0.2\linewidth]{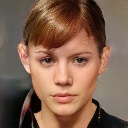}\includegraphics[width=0.2\linewidth]{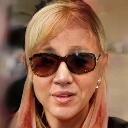}
  \includegraphics[width=0.2\linewidth]{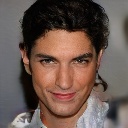}\includegraphics[width=0.2\linewidth]{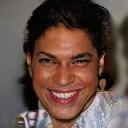}\includegraphics[width=0.2\linewidth]{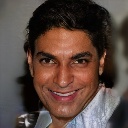}\includegraphics[width=0.2\linewidth]{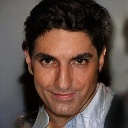}\includegraphics[width=0.2\linewidth]{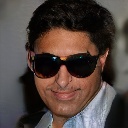}
  \caption{Two sets of synthetic samples generated using the InterFaceGAN method \cite{shen2020interpreting} with random pose, expression and eyewear.}
  \label{fig:InterFaceGAN_z_samples}
\end{figure}

\subsection{InterFaceGAN}\label{Sec. InterFaceGAN}

InterFaceGAN \cite{shen2020interpreting} is not a GAN architecture but rather a method of controlling properties in images generated by existing, pre-trained GANs. By observing generated images $G(\mathbf{z}; \theta)$ and associating binary attribute labels with latent vectors, $\mathbf{z}$, e.g. ``smiling/not smiling'', classifiers can be trained to find the hyper-planes in the latent space separating these image characteristics. An image characteristic can then be controlled by traversing the latent space along the axis perpendicular to the associated hyper-plane. Image sets were generated using the publicly available version of the Progressive GAN trained on CelebA-HQ \cite{karras2018progressive}. A thousand random $\mathbf{z}$ were first selected and then ten different images generated for each by traversing the latent space by random distances in the pose, smile and eyewear directions. Final distances from each of the three hyper-planes were selected from standard Gaussian distributions. We then scaled the pose distances by $0.15$ to give the same standard deviation of yaw angles as for the SD-GAN. In \cite{shen2020interpreting} correlations were observed between eyewear and gender. To help avoid these unwanted changes to the identity, the eyewear boundary conditioned on gender, provided by the authors, was used. Two sets of samples generated by PGAN using the InterFaceGAN method can be seen in Figure \ref{fig:InterFaceGAN_z_samples}.

\begin{table*}[t]
\begin{center}
\begin{tabular}{|l|cccc|c|}
\hline
Dataset & FRR@$3677.5$ & FRR@FAR=$10^{-4}$ & StdDev(Yaw) & LPIPS-Intra & LPIPS-Inter \\
\hline\hline
Mugshots & $2.97\times10^{-3}$ & $2.97\times10^{-3}$ & $5.5^{\circ}$ & 0.402 & 0.506 \\
\hline\hline
\rowcolor{Gainsboro}
IVI-GAN (CelebA) & $4.14\times10^{-2}$ & $4.35\times10^{-1}$ & $10.1^{\circ}$ & 0.334 & 0.558 \\
\rowcolor{Gainsboro}
InterFacePGAN (CelebA-HQ) & $2.75\times10^{-1}$ & $3.78\times10^{-1}$ & $10.1^{\circ}$ & 0.315 & 0.583 \\
\hline\hline
IVI-GAN (CelebA) & $2.45\times10^{-2}$ & $2.95\times10^{-1}$ & $3.4^{\circ}$ & 0.225 & 0.527 \\
InterFaceGAN (CelebA-HQ) & $1.05\times10^{-1}$ & $1.71\times10^{-1}$ & $3.4^{\circ}$ & 0.176 & 0.555 \\
SD-GAN (Mugshots) & $4.37\times10^{-2}$ & $8.94\times10^{-2}$ & $3.4^{\circ}$ & 0.320 & 0.419 \\
SD-Triplet (Mugshots) & $1.86\times10^{-2}$ & $6.69\times10^{-2}$ & $3.4^{\circ}$ & 0.306 & 0.428 \\
\hline
\end{tabular}
\end{center}
\caption{A selection of statistics for mated image sets from various datasets. Also reported are mean, inter-class LPIPS distances. The grey rows show statistics for datasets with unscaled poses. These do not correspond to the distributions shown in Figure \ref{fig:05_IVI_FRR}.}
\label{tab:FRRs and LPIPS}
\end{table*}


\subsection{Comparison of matching-score distributions for disentangled, synthetic datasets}\label{Sec. Disentangled comparison}

\begin{figure}
  \centering
  \includegraphics[width=1.0\linewidth]{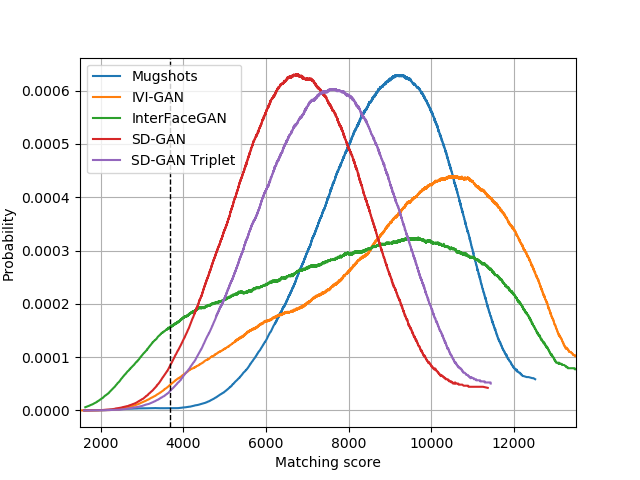}
  \caption{Distributions of matching scores for mated image sets within the real or synthetic datasets indicated in the legend.}
  \label{fig:05_IVI_FRR}
\end{figure}

Figure \ref{fig:05_IVI_FRR} shows distributions of matching scores for all mated image pairings within the various datasets. Ideally, score distributions should be similar to those calculated for the dataset of mugshots (blue curve). We note, however, that significant portions of each synthetic distribution lie below the threshold of $3677.5$. Table \ref{tab:FRRs and LPIPS} reports FRRs at this threshold, and also at points corresponding to FAR=$10^{-4}$ based on the matching scores between non-mated pairs. We have taken special care to ensure that the distributions of yaw angles are similar for each synthetic dataset. To give an idea of the remaining discrepancies in non-identity variation, Table \ref{tab:FRRs and LPIPS} also reports average perceptual distances between mated pairs (``LPIPS-Intra'') and non-mated pairs (``LPIPS-Inter'') as measured by the LPIPS perceptual similarity metric \cite{zhang2018unreasonable}. We used version 0.1 of LPIPS with a VGGNet base and additional linear calibration layer. (Note that these distances are not insensitive to changes in identity.)

\begin{figure}[t]
    \centering
    \includegraphics[width=1.0\linewidth]{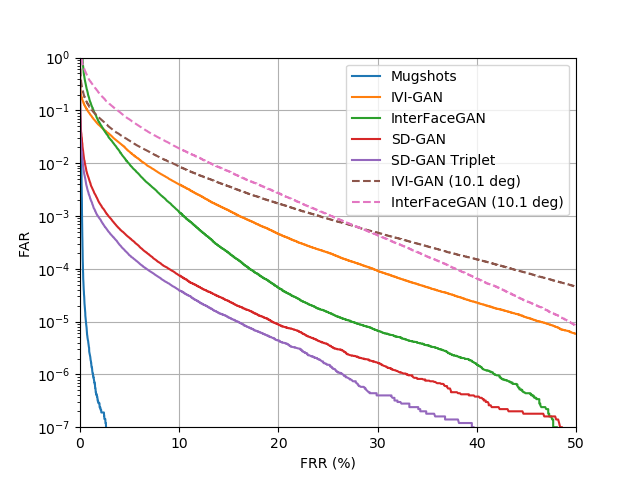}
    \caption{ROC curves for the identity-disentangled datasets.}
    \label{fig:ROCs}
\end{figure}

In absolute terms (based on FRR@$3677.5$), InterFaceGAN proved to be the least effective method at maintaining identity with an FRR of $10.5\%$ despite demonstrating lower intra-class LPIPS distances than all other datasets. It appears that, without applying an explicit biometric constraint, identity is not well disentangled from other properties in the latent space of the Progressive GAN. SD-GAN with triplet loss was found to be the most effective at preserving identity, followed by IVI-GAN and the standard SD-GAN. FRR@$3677.5$ for IVI-GAN remains lower than for the standard SD-GAN even if we do not limit the pose variation, which results in larger intra-class LPIPS distances than for SD-GAN. (See the grey rows of Table \ref{tab:FRRs and LPIPS}.) While the identity constraint of IVI-GAN is effective, using it appears to come at a cost. IVI-GAN suffers from significant mode-collapse in the biometric feature space resulting in non-mated identity collisions and higher values of FRR@FAR=$10^{-4}$ despite its relatively low values of FRR@$3677.5$. Interestingly, this collapse appears to be confined to the biometric feature space and does not strongly affect the inter-class LPIPS distances. The best compromise between consistency of identity within mated sets and variety of identity between sets is achieved by our SD-GAN with triplet loss with a value of FRR@FAR=$10^{-4}$ of $6.69 \times 10^{-2}$. It can be seen from Figures \ref{fig:CelebA_HQ_FARs_comp} (right) and \ref{fig:CelebA_HQ_FARs_NNs_comp} (right) that training using the triplet loss slightly increases mode-collapse of identity. However, the improvement in intra-class compactness compensates for this and, as can be seen from the ROC curves in Figure \ref{fig:ROCs}, disentanglement is improved at all thresholds.


\section{Conclusions}\label{Sec. Conclusions}

We performed an analysis of the ability of GANs to generate new identities. In doing so, we introduced a technique of analysing both the degree of overfitting of generators to the training dataset, and the degree of mode-collapse. We used this technique to show that overfitting is minimal and that GANs trained on face images are capable of generating new identities. This validates the use of GANs for data anonymisation and allows us to publish the synthetic identities shown in Figures \ref{fig:B61_samples} and \ref{fig:Triplet_samples} despite having used a proprietary dataset for training. It also validates the assumptions made during semi-supervised learning, for example, that synthetic images can indeed be safely assigned to a $k+1^{th}$ class without detrimentally affecting existing classes.

We also assessed the performance of methods designed to disentangle identity from other image properties. We evaluated InterFaceGAN, IVI-GAN and SD-GAN, and showed that our novel GAN triplet loss can be used to improve the disentanglement of identity. None of the algorithms, however, is able to do so to a satisfactory degree. Even the lowest value of FRR@FAR=$10^{-4}$, found for our SD-GAN with triplet loss, is more than an order of magnitude larger than that found for real data. More advanced forms of data-augmentation, involving the generation of sets of images for new, synthetic identities, are therefore unlikely to be fruitful when using methods similar to those evaluated here.

{\small
\bibliographystyle{ieee}
\bibliography{BioGAN}
}

\end{document}